\documentclass[10pt,twocolumn,letterpaper]{article}

\usepackage{multirow}
\usepackage{cvpr}
\usepackage{times}
\usepackage{epsfig}
\usepackage{graphicx}
\usepackage{amsmath}
\usepackage{amssymb}
\usepackage[export]{adjustbox}
\usepackage[numbers,sort]{natbib} 

\usepackage{multido}
\usepackage{subcaption}
\usepackage{comment}

\usepackage[pagebackref=true,breaklinks=true,letterpaper=true,colorlinks,bookmarks=false]{hyperref}

\cvprfinalcopy

\ifcvprfinal\fi
\begin{document}

\title{A Short Note on the  Kinetics-700-2020 Human Action Dataset}

\author{Lucas Smaira \\
{\tt\small lsmaira@google.com} \\
\and
Jo{\~a}o Carreira \\
{\tt\small joaoluis@google.com} \\
\and
Eric Noland \\
{\tt\small enoland@google.com} \\
\and
Ellen Clancy \\
{\tt\small clancye@google.com} \\
\and
Amy Wu \\
{\tt\small amybwu@google.com} \\
\and
Andrew Zisserman \\
{\tt\small zisserman@google.com} \\
}

\maketitle

\begin{abstract}

We describe the 2020 edition of the DeepMind Kinetics human action dataset, which replenishes and extends the Kinetics-700 dataset. In this new version, there are at least 700 video clips from different YouTube videos for each of the 700 classes. This paper details the changes introduced for this new release of the dataset and includes a comprehensive set of statistics as well as baseline results using the I3D network.

\end{abstract}

\section{Introduction}
The  Kinetics datasets are a series of large scale curated datasets of video clips, covering a diverse range of human actions.
They can be used for training and exploring neural network architectures for modelling human actions in video.

Three editions have been released: Kinetics-400~\cite{kay2017kinetics}, Kinetics-600~\cite{kinetics600} and Kinetics-700~\cite{kinetics700}, with 400, 600 and 700 human action classes, respectively.
In each case: (i) the clips are from YouTube videos, last 10s, and have a variable resolution and frame rate; and (ii) for an action class, all clips are from different YouTube videos. The statistics of the datasets are given in 
table~\ref{tab:statistics}.

\begin{table}
\centering
\begin{tabular}{| c | c | c | c |}
  \hline
  \textbf{Dataset} & \textbf{\# classes} & \textbf{Average} & \textbf{Minimum} \\ \hline
Kinetics-400 & 400  & 683 & 303 \\ \hline
Kinetics-600 & 600  & 762 & 519 \\ \hline
Kinetics-700 & 700 & 906 & 532 \\ \hline
Kinetics-700-2020  & 700 & 926 & 705 \\ \hline

\end{tabular} 
\vspace{5pt}
\caption{Statistics on the number of video clips per class for different Kinetics datasets as of 14-10-2020.}
\label{tab:statistics}
\end{table}

Building datasets of realistic videos from YouTube presents the challenge of dealing with video disappearance -- for example, due to users removing the videos or making them private. The scale of this problem is illustrated in table~\ref{tab:wipeout}.
To address this problem, we have released a new edition of the Kinetics-700 dataset, called Kinetics-700-2020, where the clips for each class have been replenished. Note, unlike in previous years we have not increased the number of classes.

\begin{table*}
\centering
\begin{tabular}{| c | r | r | r |}
  \hline
  \textbf{Dataset \& split} & \textbf{\# clips} & \textbf{\# clips  14-10-2020} & \textbf{\% retained} \\ \hline
Kinetics-400 train & 246,245 & 220,033 & 89\% \\ \hline
Kinetics-400 val & 20,000 & 18,059 & 90\% \\ \hline
Kinetics-400 test & 40,000 & 35,400 & 89\% \\ \hline\hline
Kinetics-600 train & 392,622 & 371,910 & 95\% \\ \hline
Kinetics-600 val & 30,000 & 28,366 & 95\% \\ \hline
Kinetics-600 test & 60,000 & 56,703 & 95\% \\ \hline\hline
Kinetics-700 train & 545,317 & 532,370 & 98\% \\ \hline
Kinetics-700 val & 35,000 & 34,056 & 97\% \\ \hline
Kinetics-700 test & 70,000 & 67,302 & 96\% \\ \hline\hline
Kinetics-700-2020 train & 545,793 & -- & -- \\ \hline
Kinetics-700-2020 val & 34,256 & -- & -- \\ \hline
Kinetics-700-2020 test  & 67,858 & -- & -- \\ \hline\hline
\end{tabular} 
\vspace{5pt}
\caption{The number of original (left) and current (right) available video clips in the various Kinetics datasets. }
\label{tab:wipeout}
\end{table*}

The URLs of the YouTube videos and temporal intervals of all the Kinetics datasets can be obtained from \url{https://deepmind.com/research/open-source/kinetics}. The link also includes additional annotations for the AVA-Kinetics~\cite{ava-kinetics} and Countix~\cite{Dwibed20} datasets.

\section{Data Collection Process \label{collection}}

The collection process follows that described in ~\cite{kinetics700} but focuses only on the rare classes. We collect new clips for the 123 rarest classes (containing less than 700 clips), topping up those until they reach at least 700  per class. This is shown in table~\ref{tab:statistics}. We also show yields for these classes in Appendix~\ref{sec:yieldrate}.

Since rare classes have a poor yield rate (proportion of candidate clips which are rated positive), we increased the number and quality of the text queries used to collect candidate YouTube video ids by techniques such as: using verbs in both infinitive and gerund format; removing stop words and articles; and using synonyms. The same procedure was carried out in all four query languages (English, French, Spanish and Portuguese). Augmenting the query space proved successful in helping to obtain more and better quality videos (with content more related to the class).

\vspace{3mm}
\noindent \textbf{Removing duplicates.}
The same clip can occur multiple times. This happens because: (i) the same video is  uploaded multiple times to YouTube; or (ii) different videos contain the same clip (e.g. compilations). This is common in instructional videos, particularly in classes such as 'pouring milk', 'tasting wine', 'vacuuming car'. In order to filter those clips from the final dataset, we cluster them and look at individual clusters gifs removing duplicates. A final filtering is also done to make sure clips belong to the correct class.

\vspace{3mm}
\noindent \textbf{Geographical diversity.}
We provide an analysis of the geographical distribution of the videos in the final dataset at the granularity of continents. The location is assigned based on where the video was uploaded from. The results are shown in table~\ref{tab:geolocation} based on the fraction of videos containing that information (around 90\%). 

\begin{table*}
\centering
\begin{tabular}{| c | c | c | c | c |}
  \hline
  \textbf{Continent} & \textbf{Kinetics-400} & \textbf{Kinetics-600} & \textbf{Kinetics-700} & \textbf{Kinetics-700-2020}  \\ \hline
Africa & 0.8\% & 0.9\% & 1.0\% & 1.0\% \\ \hline
Asia & 11.8\% & 11.3\% & 11.5\% & 11.7\% \\ \hline
Europe & 21.4\% & 19.3\% & 19.6\% & 19.5\% \\ \hline
Latin America & 3.4\% & 5.7\% & 7.6\% & 7.7\% \\ \hline
North America & 59.0\% & 59.1\% & 56.8\% & 56.6\% \\ \hline
Oceania & 0.8\% & 3.7\% & 3.5\% & 3.5\% \\ \hline
\end{tabular} 
\vspace{5pt}
\caption{Geographical data distribution, per continent.}
\label{tab:geolocation}
\end{table*}

Geographical diversity increased slightly over the years, especially the percentage of videos from Latin America, probably because we started querying for videos in Portuguese in the Kinetics-600 edition and also Spanish in the Kinetics-700 edition.
The multiple language queries were introduced to increase diversity and yield.
Overall, still more than half of the videos were uploaded from North America, possibly because of querying in English from the start (with Kinetics-400) but maybe also due to the greater popularity of YouTube in North America.

\section{Benchmark Performance}

As a baseline model we used I3D \cite{Carreira17}, with standard RGB videos as input (no optical flow). 
We trained the model from scratch on the Kinetics-700-2020 training set using different numbers of training examples: 100, 200, 300, 400, 500, 600 and all (some classes have up to 1000 training examples). We report performance on the validation and test sets. Results are shown in table~\ref{tab:benchmark}.

\begin{table}
\centering
\begin{tabular}{| c| r | r | c |}
  \hline
  \textbf{\# train examples} & \textbf{Valid} & \textbf{Test}  \\ \hline
100 & 38.8 / 63.0 & 36.9 / 61.1 \\ \hline
200 & 48.6 / 72.4 & 46.8 / 70.9 \\ \hline
300 & 52.4 / 76.0 & 50.8 / 74.6 \\ \hline
400 & 54.1 / 77.6 & 52.6 / 76.0 \\ \hline
500 & 55.7 / 79.0 & 54.0 / 77.7 \\ \hline
600 & 58.1 / 81.1 & 56.8 / 79.9 \\ \hline
Kinetics-700-2020 & 59.3 / 82.0 & 58.2 / 80.9 \\ \hline\hline
Kinetics-700 & 58.0 / 81.7 & 57.6 / 80.7 \\ \hline
\end{tabular} 
\vspace{5pt}
\caption{Performance of an I3D model with RGB inputs on the Kinetics-700-2020 dataset valid and test set using different number of training examples and evaluating in 8 regularly spaced clips. Each row shows top-1 / top-5 accuracy in percentage.}
\label{tab:benchmark}
\end{table}

\begin{figure}
  \includegraphics[width=\linewidth]{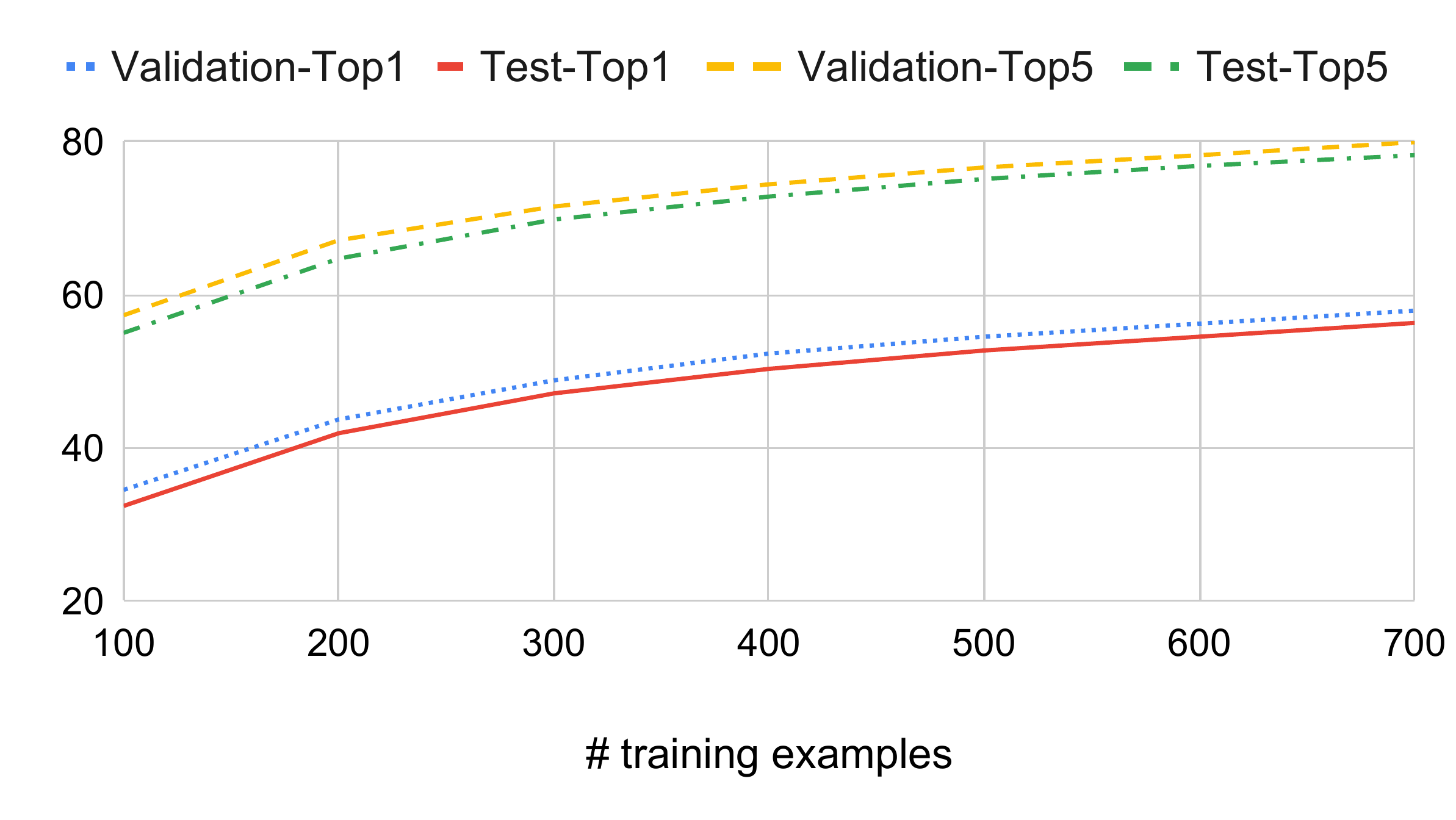}
  \caption{Performance of an I3D model with RGB inputs on the Kinetics-700-2020 dataset using different number of training examples and and evaluating  using 8 linearly spaced segments per clip.}
  \label{fig:boat1}
\end{figure}

Top-1 and top-5 accuracy improve steadily with more examples per class, even given that I3D is a model with few parameters: around 12M. In contrast, for example, a ResNet-50 model~\cite{he2016deep} has nearly double the parameters at 23M.

\paragraph{Implementation details.} We used a 32 device TPU pod, batch size 8 videos per device, 32 frame clips for training and 8 clips of 32 frames for testing. We trained using SGD with momentum set to $0.9$ and weight decay of $1e-5$ for 135 epochs. We start with a learning rate of 0.5, decreasing it by a factor of 10 after 90, 105, 115 and 120 epochs.

\section{Conclusion}
\label{sec:conclusion}

We have described the new Kinetics-700-2020 dataset, which in terms of clip counts is considerably more balanced than the current Kinetics-700 with all classes now having a minimum of 700 examples. We have also demonstrated the benefits of having more training clips in improving I3D classification performance.
The Kinetics datasets were originally introduced to aid architectural development for spatio-temporal models, and for model pre-training for downstream tasks.  With the evolution of the field and improvements in self-supervised learning, Kinetics may eventually become, in turn,  a good downstream task itself. 

\subsection*{Acknowledgements:} The collection of this dataset was funded by DeepMind.

{\small
\bibliographystyle{ieee}
\bibliography{references}
}

\appendix

\section{Yield success rate per class}
\label{sec:yieldrate}
This is the ranked list of classes to which new clips have been added, where the  first number is the probability that a candidate clip was voted positive for that class by three or more human annotators and the second number indicates a probability that an example is published in the dataset, after deduplication and the final filtering.

\begin{enumerate}
\itemsep0em 
\item stacking dice	38.68\%	37.93\%
\item steering car	65.90\%	35.44\%
\item putting on sari	48.46\%	32.54\%
\item punching person (boxing)	30.81\%	30.81\%
\item steer roping	38.29\%	30.79\%
\item making slime	37.83\%	27.71\%
\item filling eyebrows	36.47\%	27.26\%
\item washing hair	34.15\%	26.83\%
\item square dancing	46.13\%	25.94\%
\item scrapbooking	35.94\%	25.46\%
\item jumping sofa	24.85\%	24.49\%
\item threading needle	30.63\%	24.32\%
\item brushing floor	31.14\%	23.06\%
\item eating nachos	33.97\%	22.73\%
\item playing with trains	46.50\%	22.72\%
\item metal detecting	25.96\%	22.12\%
\item using atm	25.19\%	21.91\%
\item grinding meat	28.40\%	20.99\%
\item base jumping	30.65\%	20.69\%
\item springboard diving	32.88\%	20.55\%
\item tie dying	23.53\%	19.61\%
\item luge	23.42\%	18.92\%
\item playing piccolo	25.99\%	18.88\%
\item sucking lolly	30.30\%	18.83\%
\item polishing furniture	24.31\%	18.78\%
\item calculating	25.21\%	18.70\%
\item looking at phone	26.76\%	18.31\%
\item chiseling wood	23.26\%	17.44\%
\item picking apples	19.75\%	17.28\%
\item swimming with sharks	25.10\%	17.19\%
\item decoupage	21.13\%	17.01\%
\item coloring in	39.26\%	16.92\%
\item poking bellybutton	17.88\%	16.56\%
\item chiseling stone	21.19\%	16.56\%
\item doing laundry	22.31\%	16.53\%
\item tiptoeing	21.99\%	16.31\%
\item waxing armpits	22.29\%	16.28\%
\item curling eyelashes	23.80\%	16.26\%
\item pulling rope (game)	17.02\%	16.13\%
\item filling cake	21.09\%	15.99\%
\item opening coconuts	16.09\%	15.93\%
\item bending back	16.46\%	15.92\%
\item sausage making	23.31\%	15.73\%
\item passing American football (in game)	18.94\%	15.61\%
\item laying stone	21.26\%	15.28\%
\item playing blackjack	22.20\%	15.07\%
\item changing gear in car	18.82\%	14.76\%
\item home roasting coffee	17.34\%	14.49\%
\item cutting cake	17.83\%	14.44\%
\item playing rounders	16.39\%	14.23\%
\item treating wood	17.59\%	13.67\%
\item vacuuming car	18.14\%	13.41\%
\item picking blueberries	17.31\%	13.03\%
\item dealing cards	15.42\%	12.98\%
\item laying decking	13.60\%	12.13\%
\item poaching eggs	15.37\%	12.04\%
\item swimming with dolphins	14.13\%	11.96\%
\item petting horse	16.60\%	11.95\%
\item lighting candle	12.43\%	11.86\%
\item taking photo	15.29\%	11.59\%
\item dyeing eyebrows	14.31\%	11.48\%
\item gospel singing in church	14.83\%	11.30\%
\item sieving	13.38\%	11.04\%
\item cutting orange	16.80\%	11.02\%
\item carving marble	15.26\%	10.96\%
\item shoot dance	12.12\%	10.92\%
\item grooming cat	17.52\%	10.86\%
\item tasting wine	11.90\%	10.71\%
\item combing hair	20.36\%	10.69\%
\item uncorking champagne	16.31\%	10.61\%
\item skiing mono	13.30\%	10.47\%
\item putting wallpaper on wall	14.29\%	10.27\%
\item scrubbing face	12.57\%	10.20\%
\item surveying	12.42\%	9.99\%
\item looking in mirror	12.60\%	9.72\%
\item mushroom foraging	11.29\%	9.68\%
\item ski ballet	9.76\%	8.62\%
\item playing road hockey	11.25\%	8.50\%
\item applying cream	8.91\%	7.70\%
\item carving wood with a knife	8.31\%	7.42\%
\item using inhaler	7.32\%	7.32\%
\item milking goat	10.78\%	7.19\%
\item assembling bicycle	7.76\%	7.13\%
\item squeezing orange	9.36\%	7.08\%
\item pulling espresso shot	7.19\%	6.90\%
\item baby waking up	8.03\%	6.80\%
\item pouring wine	9.42\%	6.75\%
\item shopping	9.53\%	6.72\%
\item seasoning food	7.58\%	6.72\%
\item adjusting glasses	8.11\%	6.68\%
\item being in zero gravity	8.58\%	6.66\%
\item blending fruit	7.05\%	6.54\%
\item mixing colours	7.48\%	6.51\%
\item spinning plates	8.03\%	6.45\%
\item ice swimming	7.25\%	6.11\%
\item doing sudoku	7.40\%	5.75\%
\item letting go of balloon	6.09\%	5.71\%
\item fixing bicycle	5.62\%	5.62\%
\item entering church	6.28\%	5.55\%
\item chasing	5.98\%	5.32\%
\item playing shuffleboard	6.35\%	5.31\%
\item playing mahjong	11.31\%	5.20\%
\item peeling banana	6.06\%	5.14\%
\item closing door	6.25\%	4.99\%
\item shredding paper	6.55\%	4.91\%
\item card stacking	6.13\%	4.90\%
\item saluting	8.59\%	4.89\%
\item capsizing	6.26\%	4.82\%
\item delivering mail	5.12\%	4.57\%
\item listening with headphones	8.69\%	4.56\%
\item tossing salad	4.85\%	4.49\%
\item pouring milk	8.12\%	4.28\%
\item playing nose flute	5.72\%	4.25\%
\item carrying weight	6.73\%	4.13\%
\item shooting off fireworks	4.67\%	4.08\%
\item answering questions	5.87\%	4.07\%
\item testifying	12.50\%	4.04\%
\item herding cattle	4.48\%	4.04\%
\item putting on shoes	5.40\%	3.84\%
\item photobombing	4.43\%	2.96\%
\item bouncing ball (not juggling)	2.87\%	2.59\%
\item coughing	2.82\%	2.11\%
\item twiddling fingers	3.74\%	2.02\%
\end{enumerate}

\end{document}